\documentclass{article}
\usepackage{arxiv}

\usepackage[utf8]{inputenc} 
\usepackage[T1]{fontenc}    
\usepackage{hyperref}       
\usepackage{url}            
\usepackage{booktabs}       
\usepackage{amsfonts}       
\usepackage{nicefrac}       
\usepackage{microtype}      
\usepackage{lipsum}
\usepackage{authblk}
\usepackage[misc]{ifsym}
\usepackage{graphicx}
\usepackage{multicol}
\usepackage{multirow}
\usepackage{amsmath}
\begin{document}

\title{TAN: Temporal Affine Network for Real-Time Left Ventricle Anatomical Structure Analysis Based on 2D Ultrasound Videos}
\author[1]{Sihong Chen}
\author[1]{Kai Ma}
\author[1]{Yefeng Zheng\thanks{yefengzheng@tencent.com}}
\affil[1]{Tencent YouTu X-Lab, Shenzhen, China}

\renewcommand\Authands{ and }

\maketitle

\begin{abstract}
With superiorities on low cost, portability, and free of radiation, echocardiogram is a widely used imaging modality for left ventricle (LV) function quantification. However, automatic LV segmentation and motion tracking is still a challenging task.
In addition to fuzzy border definition, low contrast, and abounding artifacts on typical ultrasound images, the shape and size of the LV change significantly in a cardiac cycle.
In this work, we propose a temporal affine network (TAN) to perform image analysis in a warped image space, where the shape and size variations due to the cardiac motion as well as other artifacts are largely compensated. Furthermore, 
we perform three frequent echocardiogram interpretation tasks simultaneously: standard cardiac plane recognition, LV landmark detection, and LV segmentation.
Instead of using three networks with one dedicating to each task, we use a multi-task network to perform three tasks simultaneously.
Since three tasks share the same encoder, the compact network improves the segmentation accuracy with more supervision.
The network is further finetuned with optical flow adjusted annotations to enhance motion coherence in the segmentation result.
Experiments on 1,714 2D echocardiographic sequences demonstrate that the proposed method achieves state-of-the-art segmentation accuracy with real-time efficiency.
\end{abstract}

\section{Introduction}

The echocardiography is a widely used imaging modality for visualizing the anatomical cardiac structure and accessing potential cardiovascular diseases, as it offers a real-time, low cost and non-invasive solution for clinical routine diagnosis. Estimation of LV boundary from enchocardiograms is recognized as one of the main methods to measure heart functions. However, conventional manual delineation of LV boundary suffers from the following issues: 1) it can be accurately labeled only by an experienced sonographer; 2) it is laborious and time-consuming; and 3) it is prone to annotator variations. Therefore, a fully automatic segmentation method is highly desirable in clinical practice. 

\begin{figure}
\begin{center}
\includegraphics[width=1.0\textwidth]{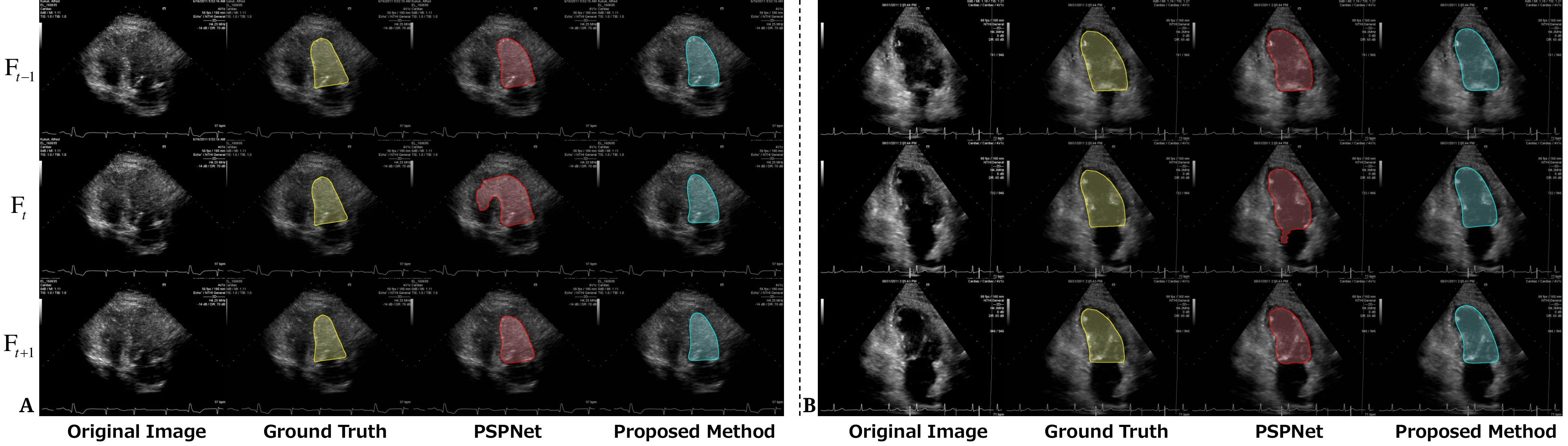}
\end{center}
   \caption{
Left ventricle segmentation results on a few challenging cases. 
(A) Fuzzy border; (B) Motion of the mitral valve leaflets.
}
\label{fig1}
\end{figure}

Automatic segmentation of LV boundary from 2D B-mode echocardiograms continues to be a challenging problem due to the facts that: 1) The size and shape of a cardiac chamber change significantly between the end-diastolic (ED) phase and the end-systolic (ES) phase; 2) Echocardiograms often suffer from the ultrasound specific artifacts. For instance, at the ES phase (when the LV is fully contracted), the ventricular wall boundary becomes blurry or even disappears (Fig.~\ref{fig1}A). When the mitral valve is fully open, the valve leaflets are merely visible (Fig.~\ref{fig1}B), making it difficult to distinguish the LV boundary from the left atrium (LA); and 3) Echocardiogram quality may differ significantly due to the specific properties and settings of ultrasound machines. Although the recent advancement of machine learning algorithms \cite{Xu1998Snakes,Zhou2010Shape,Shelhamer2014Fully,pace2018iterative} has significantly improved the segmentation results, the aforementioned challenges are not completely addressed.

In this work, we propose an automatic segmentation framework to tackle the LV analysis problem in the temporal-spatial domain. We design a novel temporal affine network (TAN) that takes a 2D echocardiographic video as the input and performs the LV segmentation in an affine warped image space, where the shape and size variations due to cardiac motions can be largely compensated. The affine warped image space of each frame is defined by three predicted landmarks on LV (apex and two mitral annulus points) from the previous frame, which preserve the semantic information (e.g., the orientation and long axis of LV). We formulate the landmark detection as an auxiliary task beyond the LV segmentation task. Moreover, we introduce another auxiliary task to recognize the cardiac plane of the ultrasound scan, e.g., the apical-2-chamber plane (A2C) vs. apical-4-chamber (A4C) plane. By sharing representations between related tasks, we can enable the proposed TAN network to generalize on the main LV segmentation task. In order to adaptively learn the motion consistency among adjacent frames, we design an end-to-end temporal coherence module (TCM) that updates the annotated LV contours by the Lucas-Kanade (LK) optical flow tracking method during training. The updated annotations are used to train TAN to generate temporal coherent segmentation. In the end, we adopt lightweight network to speed up the segmentation which brings more clinical values as the interactive acquisition of echocardiograms prefers real-time performance.  

\section{Related Work}
We briefly review the related works in three aspects: traditional medical image analysis methods, image-based deep learning methods, and video-based segmentation methods, focusing on cardiac chamber segmentation and motion tracking.

Active contours and deformable models are popular among traditional medical image segmentation methods~\cite{Xu1998Snakes,Malladi1995Shape,Ballard1982Computer,Chan2001Active,Li1995Markov,Zhu1996Region}. 
Such methods may be effective in certain applications, but face several shortcomings.
For example, they highly rely on an initial contour, which need to be close to the real contour. Nearby confounding boundaries may lead the evolving contours to converge to a wrong segmentation result. Methods utilizing texture information are more inclined to over-segmentation results~\cite{Ting2008Semiautomated}.

Recently, segmentation methods based on deep learning, mainly 2D convolutional neural networks (CNNs), become more and more popular.
Studies like \cite{Shelhamer2014Fully,Garcia2017A,ronneberger2015u,Zhen2015Direct,Chen2016Iterative,Carneiro2012The,dangi2018left,Tarroni2018A,Dong2018VoxelAtlasGAN,Vigneault2018, Dong2018VoxelAtlasGAN, Vigneault2018, pace2018iterative} can be adopted to train a network to learn the shape and appearance variations from an annotated training dataset, which improves the accuracy and robustness of segmentation models. 
In general, a deep network may improve the accuracy to a certain degree, but at a cost of excessive computations and memory consumption. Moreover, applying these segmentation methods on each frame of an ultrasound sequence separately is suboptimal since the temporal information is completely ignored.

Video segmentation methods exploit information from neighboring frames to improve the segmentation accuracy and motion coherence~\cite{Savioli2018Automated,Yan2018Left}. 
Ni et al. \cite{Savioli2018Automated} proposed to use a recurrent neural network (RNN) to track the LV motion, but RNN is more complicated and requires more training data than CNN.
Wen et al. \cite{Yan2018Left} proposed to use optical flow to enforce motion coherence in the tracking results.
However, the optical flow calculation of a full ultrasound image is time consuming and often not robust to imaging artifacts as well as fuzzy boundaries.
Instead of applying optical flow in the test phase, we use it in the training phase to refine the network, and it does not incur any overhead cost during motion tracking.

Since the shape and size of the LV change significantly within a heartbeat, we perform segmentation in a warped image subspace to improve the robustness.
From this aspect, our work is related to the spatial transformer network (STN)~\cite{Jaderberg2015Spatial}, which allows to search for the optimal transformation parameters that minimize the loss function.
However, without ground truth transformation, we cannot control the transformation matrix and the training of STN may converge to a degenerated spatial transformer matrix.

\begin{figure*}
\begin{center}
\includegraphics[width=1\textwidth]{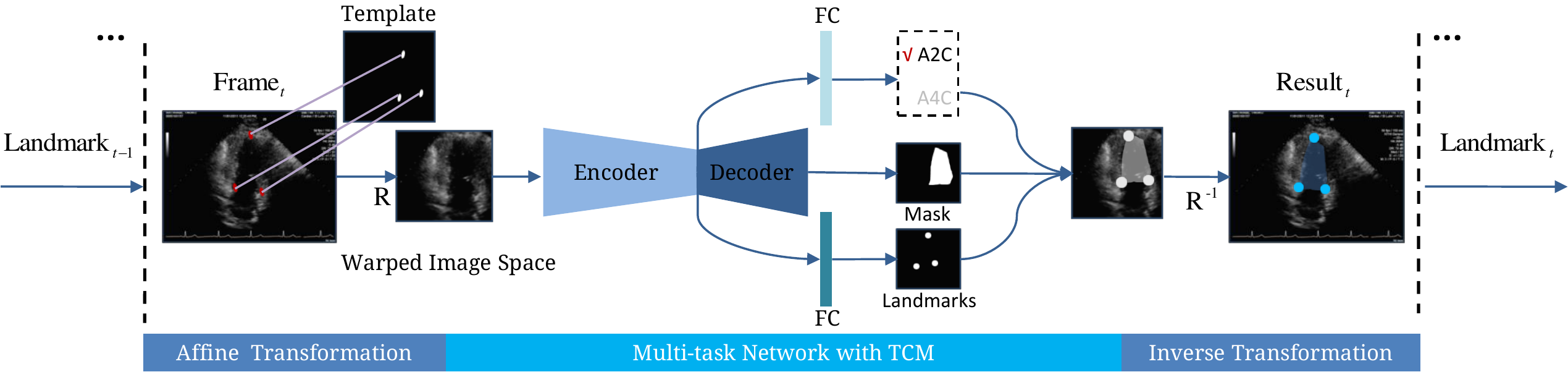}
\end{center}
   \caption{The workflow of the proposed temporal affine network (TAN) processing one echocardiographic frame.}
\label{fig2}
\end{figure*}

\section{Method}
The LV boundary often gets lost due to limited field of view, ultrasound signal dropout, and/or cardiac congestions. In the challenging cases, even experienced sonographers have to refer a couple of adjacent frames to determine the LV boundary. This is due to the fact that the LV in one frame has a similar size and shape as the one in adjacent frames, and the observation from one frame can be propagated to nearby frames as the prior information. Based on the temporal similarity observation, we design the segmentation network TAN that can incorporate the temporal information for the analysis. 
In the rest of the section, we introduce the implementation details of TAN.

\subsection{Temporal Affine Network}
As illustrated in Fig.~\ref{fig2}, TAN processes the input video frame by frame in three major steps. Firstly, the input image is warped to a canonical sub-space based on the affine transformation, which is calculated between the three landmarks propagated from the previous frame and the template landmarks derived from the training data. To be noted, we skip the warping operation of the first frame and direct estimate the landmark locations on the original image. Secondly, the warped image is analyzed by the established neural network, which simultaneously predicts the LV segmentation mask, landmark positions of the current frame and the cardiac plane category. Lastly, the updated landmarks and segmentation mask are warped back to the original image space as the final output of the current frame. Overall, the prior knowledge of the LV is propagated from one frame to the next to help improve the segmentation. The details of each sub-module are explained in the following:

\textbf{Prior shape template}. 
Three LV landmarks (i.e., the LV apex and two mitral valve annulus points) are manually annotated for each video frame in the training set. The landmark template is created by averaging the coordinates of each landmark across all video frames. Despite the differences in body shapes and visceral development of different individuals, the relative position of these three landmarks is consistent on the apical views of echocardiogram sequence. Therefore, the mean shape template can be used to encode the prior shape structure. Although the three landmarks have already defined the region of interest (ROI) of the LV, we expand the ROI boundary to cover additional regions that offer more context information for the segmentation task as well as the auxiliary ones. Since the LV always locates at the top-right position in the image, relative to other chambers in the apical view, we expand the ROI 50\% to the bottom and 50\% to the left. The expanded ROI is then normalized to a fixed size ($224 \times 224$ pixels) window, where the mean shape template is then extracted.

\textbf{Subspace image normalized from affine transformation}. A naive approach to leverage the temporal information in a video segmentation task is to propagate the segmentation ROI between adjacent frames, which offers weak location information that constrains the operating region of the segmentation. However, we believe that the temporal information shared across adjacent frames could be much more than just location information. In our workflow, we consider not only the location and range information but also the orientation and structure characteristics, which can be normalized by an affine transformation. 
Given the three landmarks from the previous frame $P_{ori}$ and corresponding ones from the shape template $P_{ref}$, we calculate the affine transformation matrix R as below:
\begin{equation}
R = F(P_{ori}, P_{ref})
\end{equation}
where $F$ denote the operation to calculate the affine transformation matrix.
Then, the original image $I_{ori}$ is warped to the canonical sub-space as:
\begin{equation}
I_{sub} = F_{R}(I_{ori})
\end{equation}
where $F_{R}$ denotes the operation of affine transformation.

In the canonical sub-space, the orientation, size and shape of the LV are more or less determined, which significantly reduces the workload of the segmentation module. Therefore, a lightweight network can be applied to achieve similar accuracy with the deeper ones, while the computation cost is much less.

\textbf{Image analysis network}. In this application, we want to perform three image analysis tasks (namely, standard plane recognition, landmark detection, and LV segmentation) and they are all performed with the warped images.
At the image analysis stage, the main task is to segment the LV region from the rest of the image. Since we utilize the temporal information by introducing additional landmarks to the workflow, another network that predicts the landmark locations is required. Though we could train two separated networks dedicating to each specific task, we design the analysis module in a multi-task fashion as the shared feature representations can help network generalize better on the main task, and are much compact to gain better runtime efficiency. In addition, a third auxiliary task that predicts class labels of the standard planes (A2C vs. A4C) is also combined into the analysis network. The logic behind this design is that the plane class is also a strong prior information for the segmentation task, which is validated by the observation that a misclassified plane often coexists with a poor segmentation result in our experiment. 

For the LV segmentation task, we compute the segmentation loss $L_{seg}$ with a pixel-wise softmax function combined with the cross entropy loss between the predicted label map and ground truth. For landmark localization task, we first normalize the x, y coordinates as:
\begin{equation}
    t_{i}^{*} = t_{i}-t_{i}^{tmpl}
\end{equation}
where $t_{i}^{tmpl}$ can be treat as the template coordinate of $i\in{x,y}$. This can be thought of as a drift of the landmark from the template.
The smooth $L_{1}$ \cite{Ren2015Faster} is adopted as the loss function. For the standard plane classification task, we calculate the two-class cross-entropy loss $L_{pln}$. By combing the three pre-defined loss functions, our final loss can be defined as:
\begin{equation}
    L = \lambda_{pln}L_{pln} +\lambda_{lmk}L_{lmk} + \lambda_{seg}L_{seg} 
\end{equation}
where $L_{pln}$, $L_{lmk}$, $L_{seg}$ denote the loss of plane classification, landmark regression and LV segmentation. And, $\lambda_{pln}$, $\lambda_{lmk}$, $\lambda_{seg}$ denote the corresponding weight coefficients, respectively.

\textbf{Inverse affine transformation}
Once the warped input image gets processed by the analytic module, the output landmarks $P_{sub}$ and segmentation mask $M_{sub}$ are transferred back to the original image space according to the inverse affine transformation:
\begin{equation} 
M_{full} = F_{R}^{-1}(M_{sub}), P_{full} = F_{R}^{-1}(P_{sub})
\end{equation}
where $P_{full}$ and $M_{full}$ denote the landmark and segmentation predictions of the original image, respectively, which will be used in the next cycle.

\begin{figure}
\begin{center}
\includegraphics[width=0.58\textwidth]{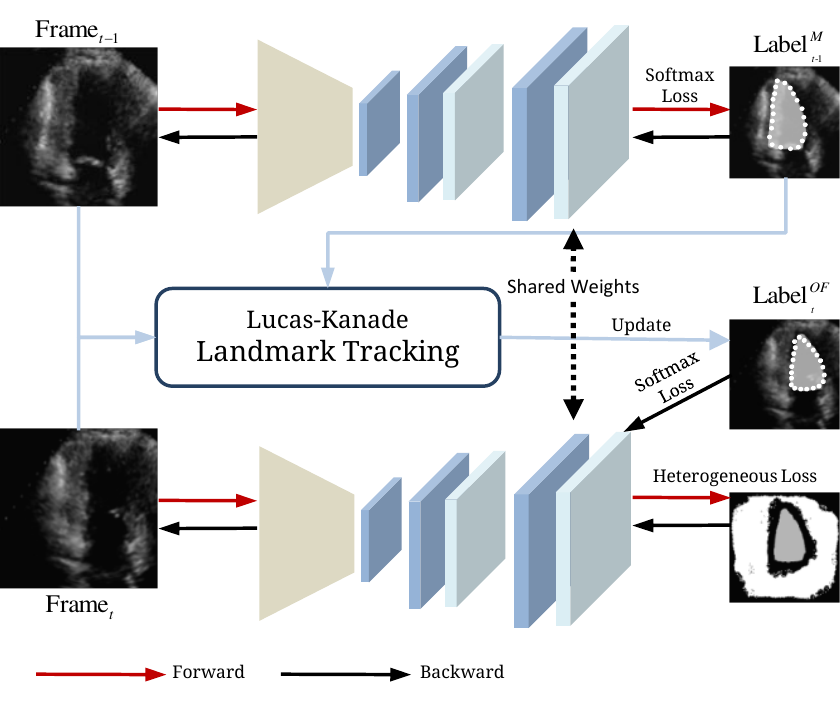}
\end{center}
   \caption{Temporal Coherence Module (TCM).}
\label{fig3}
\end{figure}

\subsection{Temporal Coherence Module (TCM)}
Since the deformation of the LV during a heartbeat cycle is progressive and continuous, temporal coherence is an important metric to measure the quality of the segmentation results. However, we find out that the segmentation annotations from the experienced sonographers sometimes violate such coherence, as they annotate the LV contour in a frame-by-frame fashion. To enforce the temporal coherence of the system while leveraging the existing annotations, we design a specific dual-path network to predict the segmentation mask of the current frame based on the Lucas-Kanade (LK) optical flow operator. As shown in Fig.~\ref{fig3} , the dual-path segmentation network is supervised by two branches of supervision during the training stage. One is from the ground truth annotation, and the other is from the generated annotation of LK optical flow tracking.

\textbf{LK operation on segmentation}. Inspired by ~\cite{dong2018supervision}, we embed the LK optical flow algorithm into the TAN framework, enforcing the model to learn the state changes between two frames. Unlike the landmark detection that has a fixed number of points as the prediction target, the segmentation contour is a continuous curve that consists of a various number of points from one frame to another. To tackle this issue, we sample a fixed number of points evenly distributed along the LV contour. In our experiment, the number of contour points is set to 20 as a tradeoff between accuracy and efficiency.
The LK optical flow equation is held for all pixels within a window centered at pixel $p_{i}$ respect to position $(x, y)$ and the local image flow (velocity) vector $(V_{x}$, $V_{y})$ must satisfy:
~\begin{equation}
    I_{x}(p_{i})V_{x} + I_{y}(p_{i})V_{y} = -I_{t}(p_{i}), i=1,2,3,...,n
\end{equation}
where $p_{i}, i={1,2,3,...,n}$ are the pixels inside the window; $I_{x}(p_{i})$, $I_{y}(p_{i})$, $I_{t}(p_{i})$ are the partial derivatives of the image $I$ with respect to $n$th window at time $t$ in the current frame. More details about LK optical flow can be found in ~\cite{lucas1981iterative}. $(V_{x}$, $V_{y})$ constitutes the transformation matrix $P$. Then, the contour points of the updated annotation can be determined by the affine transformation matrix and the points from the previous frame annotation, as illustrated in Fig.~\ref{fig3}. Finally, we connect the updated points and smooth the contour to obtain the label mask.

\textbf{Dual-path network}. 
Theoretically we can ask physicians to offline check and correct the adjusted annotations from the LK optical flow tracking operation. However, this process is not practical as it is time-consuming and not much less than re-annotating the data. Instead, we propose a dual-path segmentation network to leverage both sets of annotations. As shown in Fig.~\ref{fig3}, the top branch is trained with previous frame $Frame_{t-1}$ and the corresponding ground truth label $Label^{M}_{t-1}$, while the bottom one is trained with the current frame $Frame_{t}$ and the updated annotation $Label^{OF}_{t}$ from the LK optical flow tracking. The two networks are trained simultaneously and the weight parameters are shared in between. During the inference stage, we extract only one path of the network to process the data and get the output.

\textbf{Heterogeneous loss with OHEM}. 
The dual-path network of TCM generates two segmentation losses, $L^{M}_{t-1}$ and $L^{OF}_{t}$,which correspond to each path, respectively. For faster training convergence, we design an additional heterogenous loss function that can help TCM select effective annotations. Due to noise and other artifacts in the input data, the tracking result of the contour points is not always correct and a failed tracking process can lead to a wrong segmentation annotation. When there is a tremendous difference between the manual annotation and the one from optical flow tracking, the network could be confused to have oscillated behavior or even converge to a wrong direction. Therefore, we apply the Hausdorff distance to measure the coincidence of $Label^{M}_{t}$ and LK optical flow tracking label $Label^{OF}_{t}$. When the Hausdorff distance is larger than a pre-defined threshold, the combined loss is not back-propagated to update the network parameters. The function of heterogeneous loss composed of two segmentation losses $L^{M}_{t}$ and $L^{OF}_{t}$ for $Frame_{t}$ is formally defined as below:
\begin{equation}
L^{Het}_{t} = \begin{cases}
L^{M}_{t-1} + L^{OF}_{t} & \text{ if } f_{Haus}(t) <= thresh \\ 
0 & \text{ others }
\end{cases}
\label{euq6}
\end{equation}
where t denotes the time index and $f_{Haus}$ presents the Hausdorff distance.

In echocardiographic videos, the content change between two adjacent frames usually happens around the LV boundary. We perform the online hard example mining (OHEM) with cross-entropy loss, to let the network pay more attention to the boundary regions. From the final loss feature map calculated by Eq. (6), OHEM selects 10\% pixels with the largest errors for the back-propagation. The cross-entropy loss based on OHEM operation is defined below:
\begin{equation}
L^{OF}_{t}=-\sum^{C}_{l=1}{y_{x,l}}log( p_{l\left ( x \right )}), x ~\in Z^2_{OHEM}
\end{equation}
where $p_{l\left ( x \right )}$ denotes the softmax activation; $l$ is the label of each pixel; $y_{x,l}$ is the binary indicator and $Z^2_{OHEM}$ represents the input image with pixels selected by OHEM.

\section{Experiments}
In this section, we first evaluate the effectiveness of the proposed TAN in semantic segmentation. Then, we show qualitative and quantitative evaluations of the segmentation results. We also perform ablation studies of TAN and TCM. Finally, we analyze the efficiency and accuracy of our method compared to state-of-the-art methods, such as PSPNet~\cite{ronneberger2015u} and U-Net~\cite{ronneberger2015u}.

\subsection{Training}
\textbf{Datasets}. We collected A2C echocardiogram sequences from 723 patients with a total number of 32,551 images and A4C sequences from 991 patients with 33,047 images, where each video sequence contains 21 to 100 frames.
The LV endocardial border of all images was manually annotated by medical experts. We randomly sample 90\% sequences as the training data and the rest 10\% as the test data. To determine hyperparameters, we perform a pre-study on the training dataset by further splitting it into 80\% for training and 20\% for validation.
Once the hyperparameters are determined, we use all training samples to train networks and evaluate them on the test dataset.

\textbf{Pre-processing}. 
As our network is trained on shape normalized images after an affine transformation, we warp all the training images according to the template landmarks and three annotated landmarks.
To prevent overfitting and improve the network robustness, we utilize three data augmentation techniques including translation, rotation and scaling. Considering that the LV usually does not drastically change its shape,, we set the range of augmentation deformation within the maximal movement of landmarks between neighboring frames (which is 20 pixels as calculated from the training set). 
The following settings were used in the random data augmentation during training: rotation within [$-5^o$, $5^o$], scaling within [0.8, 1.2] times of the initial image size, and translation in any direction no more than 14 pixels.

\begin{table}
\begin{center}
   \caption{Details of decoder branches for LV segmentation, landmark detection, and cardiac plane classification where deconv, conv and fc present deconvolution layers, convolution layers and fully-connected layers, respectively.}
\begin{tabular}{c|c}
\hline
Segmentation & Detection \\ \hline
\multirow{8}{*}{\begin{tabular}[c]{@{}c@{}}Deconv1 2x2, 160 stride 2\\ Conv1 3x3, 160 x2\\ Deconv2 2x2, 96 stride 2\\ Conv2 3x3, 96 x2\\ Deconv3 2x2, 64 stride 2\\ Conv3 3x3, 64 x2\\ Deconv4 2x2, 32 stride 2\\ Conv4 3x3, 32 x2\end{tabular}} & \multirow{3}{*}{\begin{tabular}[c]{@{}c@{}}FC1 256\\ FC2 6\end{tabular}} \\
 &  \\
 &  \\ \cline{2-2} 
 & Classification \\ \cline{2-2} 
 & \multirow{4}{*}{\begin{tabular}[c]{@{}c@{}}FC1 256\\ FC2 6\end{tabular}} \\
 &  \\
 &  \\
 &  \\ \hline
\end{tabular}
\label{table1}
\end{center}
\end{table}

\textbf{Network}. For the efficiency purpose, we adopt a lightweight backbone network, MobileNetV2~\cite{Sandler2018MobileNetV2}.
The encoder part downsamples the input 32 times, and then feed it to three task-specific branches for plane classification, landmark detection, and LV segmentation, respectively.  
For the plane classification and landmark detection branches, we add a fully-connected layer to each of them (see Table~\ref{table1}). For the LV segmentation branch, the decoder part is composed of four sets of upsampling modules (see Table \ref{table1} for details). In our experiments, different weight coefficients are used for different tasks as these tasks are not equally important. We empirically set the task importance weights $\lambda_{pln}$, $\lambda_{lmk}$, $\lambda_{seg}$ to 0.1, 0.1, 10.0, respectively, which achieves a good overall result. In addition, we empirically set the threshold used to measure the coincidence of $Label^{M}_{t}$ and $Label^{OF}_{t}$ to 20.0.  
\subsection{Evaluation Metrics}
The Dice coefficient and average surface distance (ASD) are used to evaluate the segmentation accuracy. The Dice coefficient evaluates the coincidence area between the segmentation prediction and ground truth, while the ASD is used to assess the fitness between two sets of contours. To further evaluate performance on motion smoothness, we also propose a smoothness index (SI) metric as follows: 
~\begin{equation}
    SI = \frac{\sum_{k=1}^{N_{v}}(\frac{1}{N_{f}^{k}-1}\sum_{t=1}^{N_{f}^{k}}ASD_{t}^{k})
}{N_{v}}
\end{equation}
where $ASD_{t}^{k}$ is the ASD between $frame_{t-1}$ and $frame_{t}$ of video $k$; $N_{v}$ denotes the number of testing videos and $N_{f}^{k}$ denotes the number of frames corresponding on the $k^{th}$ video. SI is a complementary metric to Dice coefficient and ASD. Suppose the contour on each frame is consistently one pixel off the groundtruth (i.e., $ASD=1$). Contours jumping across the groundtruth contours back and forth (i.e., jittering) result in a higher SI than contours never jump across. We also evaluate the plane classification task with the accuracy (acc) metric. As for the LV landmark detection task, following \cite{wu2017ai}, the object keypoint similarity (OKS) is used to calculate the similarity between different landmarks.

\begin{table}
\begin{center}
   \caption{Quantitative evaluation of the LV segmentation accuracy with various network encoder architectures.}
\begin{tabular}{c|c|c}
      \hline
      Backbone & Dice & ASD \\ \hline\hline
       U-Net & 88.42\% & 11.4 \\ \hline
       U-Net (TAN) & 89.97\% & 9.7 \\ \hline\hline
       PSPNet & 90.50\% & 7.1 \\ \hline
       PSPNet (TAN) & \textbf{91.11\%} & \textbf{7.0} \\ \hline\hline
       MobileNetV2 & 82.07\% & 10.7 \\ \hline
       MobileNetV2 (Crop) & 85.51\% & 10.5 \\ \hline
       MobileNetV2 (TAN) & 90.47\% &  7.1 \\ \hline
   \end{tabular}
\label{table2}
\end{center}
\end{table}

\begin{figure}
\begin{center}
\includegraphics[width=1.0\textwidth]{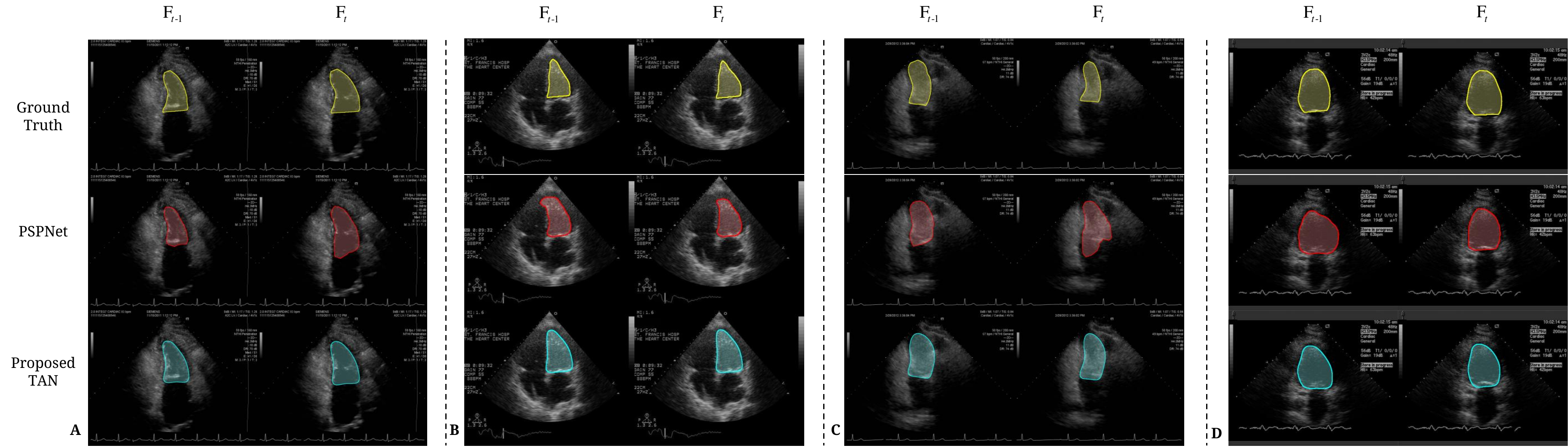}
\end{center}
   \caption{LV segmentation results using PSPNet and proposed TAN (with MobileNetV2 as the backbone).}
\label{fig4}
\end{figure}

\subsection{Effectiveness of the proposed TAN method}
To validate the effectiveness of TAN, we design three sets of comparative experiments based on the original MobileNetV2, U-Net and PSPNet, respectively, for the main segmentation branch, without the TCM dual-path architecture. Since LV segmentation is the main task of this work, we only report the segmentation accuracy in Table~\ref{table2}. The segmentation results with/without the TAN structure are compared in each set of experiments. Furthermore, in order to separately validate how much improvement comes from the utilization of temporal information and the portion from the shape normalization, we also compare with a basic cropping method that just propagates the ROI information without shape normalization.

As shown in Table ~\ref{table2}, without TAN, MobielNetV2 is significantly worse than U-Net and PSPNet because it lacks capacity to handle severe imaging artifacts and large shape variations of the LV in our dataset.
By leveraging temporal information and shape normalization, TAN significantly increases the segmentation accuracy of MobileNetV2 from 82.07\% to 90.64\% in Dice coefficient, outperforming U-Net and PSPNet without TAN.
The cropping method achieves 3.44\% matching improvement in Dice coefficient compared to the baseline MobileNetV2. Since TAN can leverage both temporal information and prior shape knowledge, it is more effective than the basic ROI propagation. The shape prior in TAN provides additional 4.96\% increase in Dice coefficient, compared to the cropping method.
TAN also improves the accuracy of U-Net and PSPNet, but the improvement is marginal.
Due to the difficulty of this task, there is a relatively large inter-observer variability and according to ~\cite{Chen2016Iterative} the agreement between annotations from two physicians ranges from 90\% to 95\% in Dice.
The room for improvement is low for large networks, however, TAN still achieves around 0.61\% and 1.55\% improvement in Dice coefficient on PSPNet and U-Net, respectively.
To conclude, with TAN, all three network architectures achieve roughly the same accuracy and MobileNetV2 is preferred since it is much faster and can meet the real-time efficiency requirement.

Fig.~\ref{fig4} shows some segmentation results of MobileNetV2+TAN and PSPNet without TAN, where TAN achieves coherent motion tracking, while the results of PSPNet are jittering. For example, as shown in the Figs.~\ref{fig4}A and ~\ref{fig4}C, when the mitral valve leaflets completely open, the boundary between the LV and the LA is hard to discern. By exploiting temporal information from neighboring frames, TAN can achieve better coherent and higher accuracy. In addition, TAN also produces more stable and accurate result around fuzzy boundaries caused by congestions or shadowing effects, as shown in Figs.~\ref{fig4}B and ~\ref{fig4}D.

\begin{figure*}
\begin{center}
\includegraphics[width=1\textwidth]{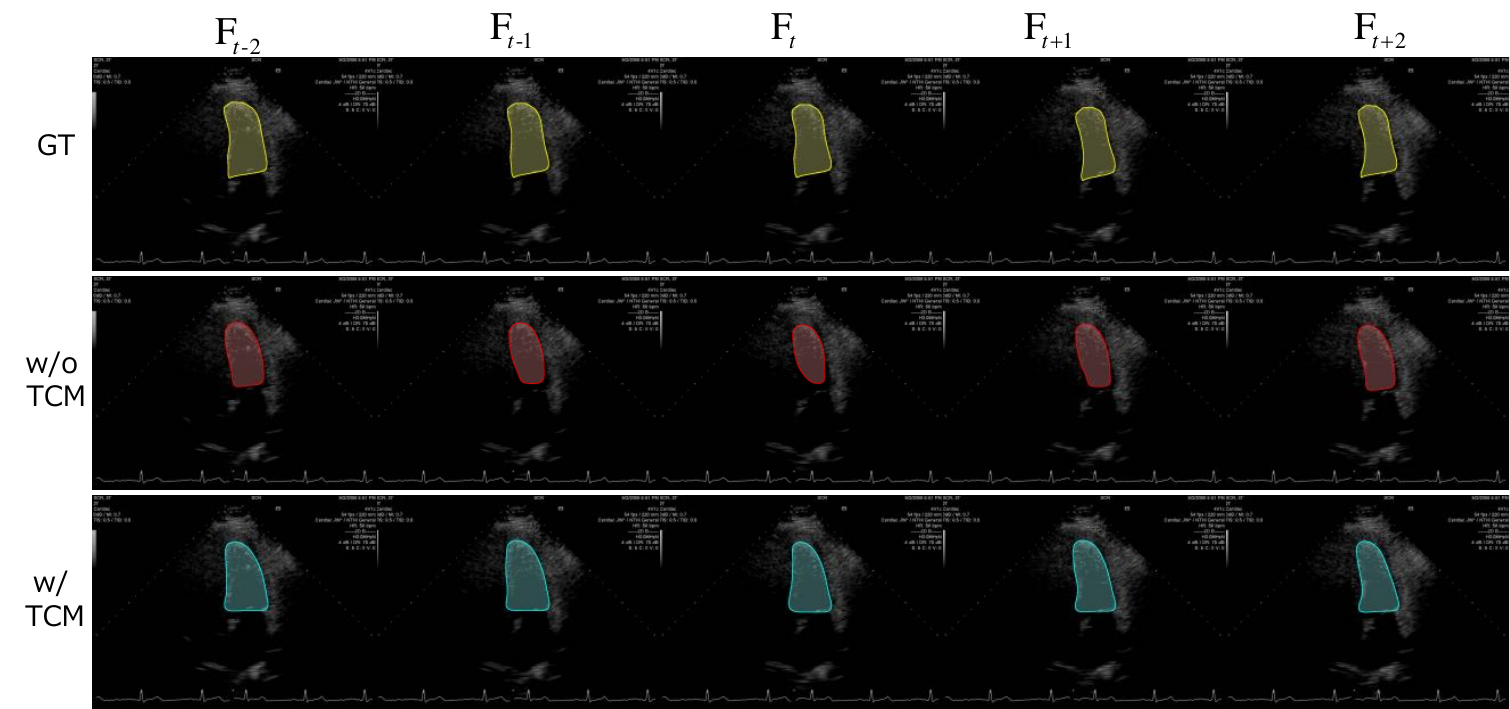}
\end{center}
   \caption{LV segmentation result without (w/o) and with (w/) the proposed temporal coherence module (TCM).}
\label{fig5}
\end{figure*}

\subsection{Effectiveness of the proposed TCM method}
\begin{table}
\begin{center}
   \caption{Quantitative evaluation of the TCM. Where “0.1” and “1.0” represent 10\% worst cases and all samples, respectively. And, ``B", ``U", ``P" and ``M" represent backbone, U-Net, PSPNet and MobileNetV2, respectively.}
\begin{tabular}{c|c|c|c|c|c|c|c}
\hline
\multirow{2}{*}{B} & \multirow{2}{*}{TCM} & \multicolumn{2}{c|}{Dice (\%)} & \multicolumn{2}{c|}{ASD} & \multicolumn{2}{c}{SI} \\ \cline{3-8} 
 &  & 0.1 & 1.0 & 0.1 & 1.0 & 0.1 & 1.0  \\ \hline\hline
U &  & 69.44 & 89.97 & 26.6 & 9.7 & 10.3 & 2.7
\\ \hline
U & \checkmark & 77.23 & 90.28 & 18.9 & 8.9 & 8.9 & 2.4
\\ \hline
P &  & 72.89 & 91.11 & 21.9 & 7.0 & 9.2 & 2.0 \\ \hline
P & \checkmark & \textbf{80.24} & \textbf{91.23} & 16.4 & 6.9 & 8.3 & 1.9 \\ \hline
M &  & 70.13 & 90.47 & 22.4 & 7.1 & 9.0 & 2.3 \\ \hline
M & \checkmark & 79.54 & 91.14 & \textbf{15.0} & \textbf{6.7} & \textbf{7.8} & \textbf{2.0}\\ \hline
\end{tabular}
\label{table3}
\end{center}
\end{table}

As a complementary part of TAN, the TCM is mainly used to adaptively fine-tune the segmentation contour of the target, making the contour more consistent from one frame to another. In TCM, we also set up three sets of comparative experiments based on MobileNetV2, PSPNet, and U-Net, respectively. The quantitative comparisons are shown in Table~\ref{table3} and all of the experiments are proceeded with the TAN framework. Since TCM is most effective on correcting failures on challenging cases, we also report error metrics on the worst 10\% cases of each algorithm.

As can be seen from Table~\ref{table3}, the performance of all networks is improved with TCM. TCM calculates the contour difference of the LV between two adjacent frames by introducing the optical flow operation, which enforces the network to pay more attention to the object boundary, so that the segmentation contour is more coherent and achieves the purpose of motion smoothing.

In Table~\ref{table3}, although TCM does not obviously improve the evaluation metrics for all samples, it gains almost 10\% increase on Dice coefficient for bad cases. More importantly, due to the effectiveness of TCM on contour prediction, ASD is greatly improved from 37\% to 48\% since the ASD is more sensitive to outliers on the contour. Similarly, we also report SI for the worst 10\% bad cases in Table~\ref{table3}.
With TCM, the SI metrics have 16\%, 22\% and 22\% improvements with U-Net, PSPNet and MobilenetV2, respectively, which indicates that TCM can maintain a good temporal coherence between adjacent frames, especially for the bad cases. This is important for medical video segmentation applications because sudden changes of the segmentation could be misdiagnosed as diseases.
As can be seen from Fig.~\ref{fig5}, mis-segmentation often occurs when the object boundary is disturbed by noise or artifacts. After using TCM, although some structural information of the image is still interfered, the result has been improved because TCM enforces the temporal coherence.

\subsection{Ablation Study}
To further validate the effectiveness of different modules proposed in our method, such as multi-task of segmentation, landmark detection and plane classification, and OHEM, we design comparison experiments as shown in Table~\ref{table4} and all experiments are proceeded without TCM. The backbone of all experiments is MobileNetV2. 

We perform segmentation, landmark detection and plane classification tasks with a single-task framework and compare the result with the one from the multi-task framework. For simplicity, we compare single-task and multi-task with the cropping method. As can be seen from the second and third rows of Table~\ref{table4}, multi-task can greatly improve the accuracy of the landmark detection and plane classification tasks but bring a small amount of improvement for the segmentation task. The comparison between rows 3 and 4 in Table~\ref{table4} shows that TAN can greatly improve the segmentation and landmark detection performance compared to the cropping method.

Through the comparison between the last two rows of Table~\ref{table4}, we can see that OHEM alone has little effect on the segmentation and landmark detections tasks. In fact, OHEM is an important complementary to TCM that shifts the network attention to LV boundaries. Together with TCM, they can significantly improve the performance and speed up the training convergence, as indicated from the last row of Table~\ref{table3}.

\begin{table}
\begin{center}
   \caption{Ablation experiments. ``Str", ``O", ``STL" and ``MTL" presents the cascade strategy, OHEM, single-task, multi-task, respectively. Dice, Acc and OKS are in unity of percentage.}
\begin{tabular}{c|c|c|c|c|c|c}
      \hline
      Str & Task & O & Dice & ASD & Acc & OKS \\ \hline\hline
      Crop & STL  &  & 85.10 & 13.6 & 87.13 & 63.25  \\ \hline
       Crop & MTL  &  & 85.51 & 10.5 & \textbf{98.25} & 68.40 \\ \hline
       TAN & MTL  &  & 90.47 & 7.1 & 97.66 & 77.53 \\ \hline
       TAN & MTL & \checkmark & \textbf{90.97} & \textbf{6.9} & 97.08 & \textbf{78.53} \\ \hline
   \end{tabular}
\label{table4}
\end{center}
\end{table}

\begin{figure}
\begin{center}
\includegraphics[width=0.5\textwidth]{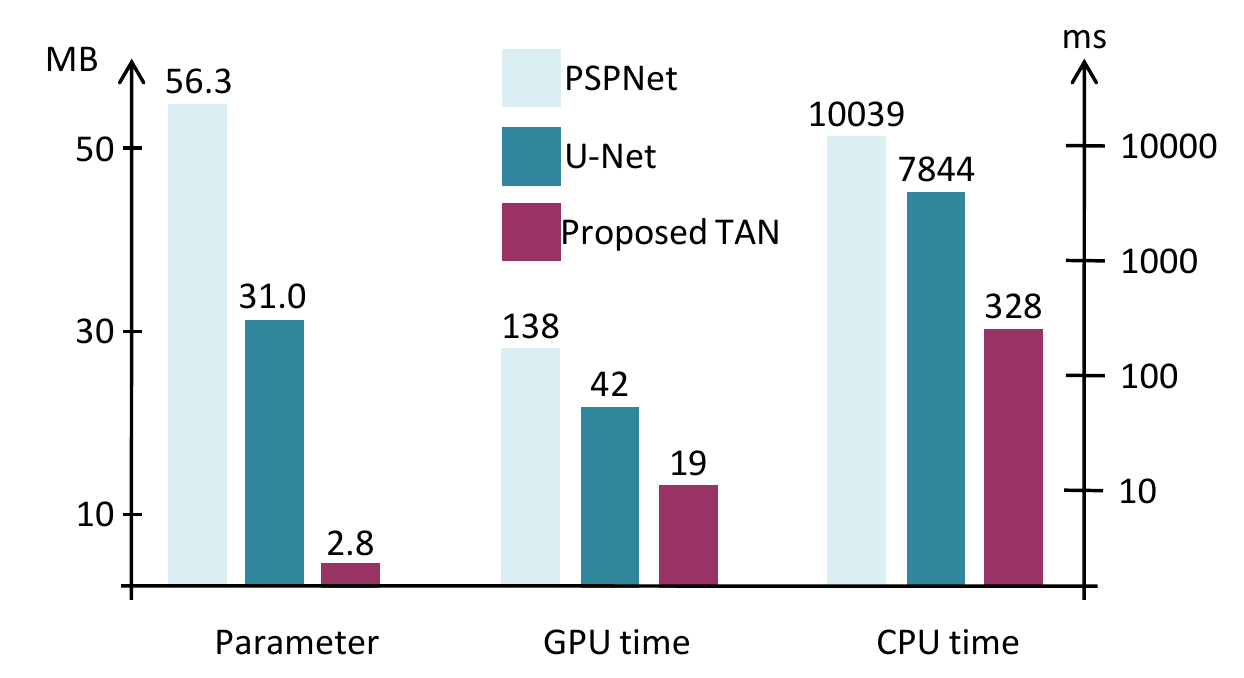}
\caption{Comparison on runtime efficiency.}
\label{fig6}
\end{center}
\end{figure}

\subsection{Runtime Efficiency}
This work aims to improve the performance of video segmentation in real-time efficiency. We calculate runtime statistics of the proposed method, PSPNet, and U-Net based on single-frame segmentation in Fig.~\ref{fig6}.
It is noted that, on a single GPU platform (NVIDIA Tesla P40), the proposed TAN can achieve real-time segmentation at 52 fps (frames per second), which is much faster than U-Net (24 fps) and PSPNet (7 fps).
Note that the latter two networks do not perform landmark detection and cardiac plane classification.
On a CPU (Intel E5-2680 v4), the runtime difference is more obvious, where TAN is more than 20 times faster than U-Net and 30 times faster than PSPNet. Although the larger networks (PSPNet and U-Net) empowered with the proposed TAN have slightly better segmentation accuracy, they are not able to achieve real-time efficiency.

\section{Conclusions}
In this work, we proposed a temporal affine network for the LV segmentation of echocardiographic video sequences. The proposed solution leveraged temporal and spatial information and greatly improved the segmentation performance to 91.14\% in Dice coefficient with a lightweight MobileNetV2 network. Meanwhile, the running-time efficiency was also improved to reach 52 fps. 

\bibliographystyle{unsrt}  

\end{document}